\documentclass[11pt]{article}

\usepackage[final]{acl}
\usepackage{nert}

\usepackage{times}
\usepackage{latexsym}

\usepackage[T1]{fontenc}

\usepackage[utf8]{inputenc}

\usepackage{microtype}

\usepackage{inconsolata}

\usepackage{graphicx}
\usepackage[skins]{tcolorbox}
\usepackage{linguex}

\usepackage{float}
\usepackage{caption} 

\newcommand{\sysname}{DeDisCo}

\title{\sysname{} at the DISRPT 2025 Shared Task: A System for Discourse Relation Classification}

\author{
    Zhuoxuan Ju,
    Jingni Wu,
    Abhishek Purushothama,
    Amir Zeldes \\
    Corpling Lab \\
    Georgetown University \\
    \{zj153, 
    jw2175,
    ap2089,
    amir.zeldes\}@georgetown.edu\\
}

\begin{document}
\maketitle
\begin{abstract}
This paper presents \textbf{\sysname{}}, Georgetown University's entry in the DISRPT 2025 shared task on discourse relation classification. We test two approaches, using an mt5-based encoder and a decoder based approach using the openly available Qwen model. We also experiment on training with augmented dataset for low-resource languages using matched data translated automatically from English, as well as using some additional linguistic features inspired by entries in previous editions of the Shared Task. Our system achieves a macro-accuracy score of 71.28, and we provide some interpretation and error analysis for our results.\footnote{Full disclosure: our team includes both organizers and
dataset annotators from the shared task. Results were reproduced independently by other organizers. All code is available at \url{https://github.com/gucorpling/disrpt25-task}.}
\end{abstract}

\section{Introduction}
Recent computational work on discourse relations has introduced richer multilingual corpora \cite{liu-etal-2024-gdtb, zeldes-etal-2025-erst} and advanced transformer-based methods for implicit and explicit discourse relation classification (e.g.~ \citealt{li-etal-2024-discourse, metheniti2024feature}). While most previous studies focus on implicit relations within a single language \cite{liu-strube-2023-annotation}, the DISRPT shared-task setting requires handling both explicit and implicit relations across multiple languages and annotation frameworks \cite{braud-etal-2024-disrpt}. This means that for example explicit \textsc{causal} relations as in \ref{ex:cause1}, where the relation can be identified thanks to the explicit connective `because', are targeted next to implicit ones, as in \ref{ex:cause2}, where it is implied that it's hard for the speaker to think about food (because) they are not hungry (both example taken from \texttt{eng.erst.gum}).

\ex. [I bought it]$_{unit1}$ [\textbf{because} it was funny]$_{unit2}$\label{ex:cause1}

\ex. [I'm so not hungry right now,]$_{unit1}$ [it's hard for me to think about food.]$_{unit2}$\label{ex:cause2}

In DISRPT 2025 (see \citealt{braud-etal-2025-disrpt}), we designed and tested two architectures -- one based on an encoder model and one based on a decoder -- and submitted the decoder-only system, which is called DeDisCo (\textbf{De}coder-based \textbf{Dis}course \textbf{Co}gnoscente), which was trained via supervised fine-tuning with instruction-style prompts, tailored for multilingual discourse relation classification. Given the shared task constraint of a single multilingual model, supervised fine-tuning allowed us to efficiently fine-tune a compact decoder architecture while maintaining robust generalization across languages. Although we also explored encoder-based alternatives for methodology comparison, the decoder model formed the core of our official submission.

\section{Related Work}

\subsection{Discourse Frameworks and Datasets}

Multiple frameworks have been used to analyze and organize  natural language discourse such as RST \cite{mann1988rhetorical} and eRST \cite{zeldes-etal-2025-erst}, PDTB \cite{marcu2004phrase}, SDRT \cite{lascarides2007segmented}, Discourse Dependencies \cite{stede2016parallel}, and ISO 24617-8 \cite{tomaszewska-etal-2024-iso}. These frameworks apply different structural and semantic guidelines for this purpose. However all of these frameworks utilize components such as discourse units and relations to build either shallow, or hierarchical/graph based discourse structures. Discourse units are then combined with \textit{discourse relations labels} to describe the complex discourse structures present in texts. 

Considerable efforts have been dedicated to building discourse corpora in multiple languages containing various text types. Combined with the variety of frameworks, many corpora contain features specific to languages, genres, and frameworks: for example, in RST, most datasets use clauses as discourse units, but only some datasets allow for discontinuous units; in PDTB, some datasets focusing on implicit relations only allow relations between sentences, while in others, smaller units are included. Additionally, the nature of the text will make some relations more or less prevalent, such as questions (the \textsc{query} relation), which appear almost only in dialogic text types. Hence, both language and genre-specific considerations can be applied in many instances, leading to heterogeneity in the data parameterized by the language, framework, corpus, and genre. Previous DISRPT Shared tasks \cite{zeldes-etal-2019-disrpt,zeldes2021disrpt,braud-etal-2023-disrpt} have brought together collections of these varied datasets, with the previous iteration \cite[DISRPT 2023][]{braud-etal-2023-disrpt} containing 26 corpora annotated across 4 frameworks and 13 languages, and the 2025 edition bringing these up to 39 datasets in coming from 6 frameworks and 16 languages. New in this edition is also the unification of the label set to the same 17 labels across all datasets; however, labels may exhibit subtle differences in usage across corpora, meaning once again that encoding dataset specific input may be crucial for high performance across benchmarks.

\subsection{Discourse Parsing and Relation Classification}

The task of discourse parsing refers to constructing discourse structures (in a given scheme) from natural language text. This involves sub-tasks such as segmenting discourse units, connecting them in order, identifying lexical units that signal relations (in PDTB and eRST), and assigning relation labels.
In addition to the inherent challenges of these tasks, heterogeneity leads to systems that are useful for only a small subset of the corpora. With neural models, this challenge is more severe due to the need for substantial amounts of training data to build such systems. The multilingual and multi-track nature of the DISRPT shared task may alleviate some of these problems by posing the problem as a multilingual one across all datasets, allowing low-resource languages to gain in performance from the additional data available in high-resource languages.
The tracks on discourse unit segmentation, connective identification, and relation classification also allow systems to focus on these problems in isolation -- in this paper we target only the latter task of relation classification, given the units connected by the relation.

Regardless of the labels used, which are commonly framework-specific \cite{hovy-1990-parsimonious}, relation classification can be considered a single-label classification task over two ordered, non-overlapping textual input units. Multiple methods have been applied to the task, including in the context of full hierarchical discourse parsing, such as shift-reduce parsing \cite{marcu-1999-decision}, feature based and neural parsers \cite{ji-eisenstein-2014-representation}, and simple span-based transformer encoder \cite{gessler-etal-2021-discodisco, metheniti2024feature}, and decoder \cite{anuranjana-2023-discoflan} models.

\subsection{Feature Encoding and Data Augmentation}

Several past top systems \citet{yu-etal-2019-gumdrop,gessler-etal-2021-discodisco,metheniti-etal-2023-discut} used hand-crafted features to mitigate discourse parsing challenges, even with transformer-based models. These range from corpus-level features such as language and framework, document-level features such as length or genre, and sample- or unit-level features such as lexical overlap or position. Unique features like whether two units share the same speaker were also crafted in corpus-specific systems. Features can be represented categorically (e.g., one-hot), as dense embeddings in neural architectures, or as plain text in LLM prompts.


Data augmentation has been used to improve both dataset and language-specific performance. Task-specific augmentation entails transforming or synthesizing data that is similar to the target dataset based on a distribution that also adheres to the dataset design. e.g. \citet{liu-etal-2023-hits} grouped together data for multiple languages with smaller corpora for such augmentation. For language-specific augmentation, the same or similar source for different languages can be translated into the target language, providing higher training data for the target language. This is popularly known as the translate-train paradigm \cite{conneau-etal-2018-xnli}, which we employ below.



\section{Data and Approach}

The Discourse Relation Classification task of the DISRPT 2025 aims to classify discourse relations across a diverse set of languages and annotation frameworks. The relation classification task includes 38 of the joint task's 39 corpora, spanning 16 languages and 6 different frameworks, with a unified set of 17 labels provided. For convenience, the language, corpora, and framework are part of the listed in \cref{sec:appendix-corpora} (\autoref{tab:corpora-list}). The majority of the data are annotated using Rhetorical Structure Theory \cite[RST;][]{mann1988rhetorical} and the Penn Discourse Treebank framework \cite[PDTB;][]{marcu2004phrase}. A smaller subset of the corpora is annotated using SDRT, discourse dependencies (DEP), eRST, and the ISO frameworks.


For our system, we experimented with both encoder-based and decoder-based models. After evaluating their performance, we selected the stronger decoder-based model for our official submission. Below, we described how each model was implemented with our feature set and data augmentation. 

\subsection{Features}
\label{sec:features}
\paragraph{Language Corpus Framework (LCF)}
As noted, the training data spans multiple annotation frameworks, corpora, and languages, which poses a challenge for generalization since relations defined under different schemes or languages are not always directly comparable. To address this diversity, we incorporate metadata into the input to help the model distinguish and generalize across datasets. We refer to these elements collectively as LCF features (Language, Corpus, Framework). For example, the dataset identifier \texttt{zho.rst.gcdt} indicates a Chinese corpus (\texttt{zho}) named GCDT \cite{peng_gcdt_2022}, annotated using the RST framework.

\paragraph{DiscoDisco Features} The DiscoDisco system \cite{gessler-etal-2021-discodisco} introduced several hand-crafted discourse features extracted from the data for the relation classification task (see \autoref{tab:discodisco_features}), which were later shown to be highly effective \cite{metheniti-etal-2023-discut}. We selectively incorporated a subset of these features into our models, see \autoref{tab:discodisco_breakdown} in \cref{sec:appendix-feature} for a detailed list of which DiscoDisco features are used in the decoder and the encoder separately. For exact details of these features we refer to the original DiscoDisco paper and system implementation.

\begin{table*}[t]
\small
\centering
\begin{tabular}{@{}lclp{6.5cm}@{}}
\toprule
\textbf{Feature} & \textbf{Type} & \textbf{Ex.} & \textbf{Description} \\
\midrule
Genre & Cat. & reddit & Document genre (e.g., eng.erst.gum) \\
Children & Num. & 2 & Child units each discourse unit has \\
Discontinuous & Cat. & false & Unit tokens not contiguous \\
Is Sentence & Cat. & true & Unit is a complete sentence \\
Length Ratio & Num. & 0.3 & Token length ratio (u1 vs. u2) \\
Same Speaker & Cat. & true & Same speaker for u1 and u2 \\
Doc. Length & Num. & 214 & Document length in tokens \\
Position & Num. & 0.4 & Unit position in doc (0–1) \\
Distance & Num. & 7 & Other units between u1 and u2 \\
Lexical Overlap & Num. & 3 & Shared non-stoplist words \\
\bottomrule
\end{tabular}
\caption{\small Features used in 2021 DiscoDisco system.}
\label{tab:discodisco_features}
\end{table*}

\paragraph{Direction} Discourse relations are marked between two segments of text, referred to as Arg1 and Arg2. In the DISRPT datasets, these segment pairs follow the text's original sequence, and an extra column specifies the intended argument direction for the annotated relation (e.g., \texttt{1>2}). For example in a cause relation, the cause points to the result, but may appear first (\texttt{1>2}) or second (\texttt{1<2}). This directional information is incorporated into both of our models, but encoded in different ways (see below).

\paragraph{Context} Text context beyond the two units being classified plays a crucial role in discourse relation classification. Notably, ~\citet{dai-huang-2018-improving} demonstrated that paragraph-level context significantly enhances the prediction of implicit discourse relations. In our models, we explore adding context to the model input and the impact of varying context window sizes surrounding the target sentence.


To clarify the experimental setup, \autoref{tab:overview_features} in \cref{sec:appendix-feature} provides an overview of which features, context, and data augmentation used in each model configuration. 

\subsection{Data Augmentation}

Given the limited size of training data for certain languages, many of which also exhibit lower model accuracy, we apply targeted data augmentation to enhance performance. We focus on six low-resource languages: Czech (14.6K tokens), Dutch (24.9K), French (32.7K), Basque (45.7K), German (66K), and Persian (67K), covering a total of seven datasets. While these six corpora are not strictly the smallest by token count, they exemplify low-resource conditions due to the combination of restricted training data for the entire language, and comparatively weak baseline performance. We therefore targeted them for augmentation to mitigate these weaknesses. Our augmentation strategy involves translating English training instances from a source corpus into the target languages using API calls to the ChatGPT 4.1 model \citep{openai2024gpt4technicalreport}, and providing this data for system training and replication following the shared task rules (our system does not access ChatGPT in any way at training or testing time). 

To ensure compatibility with the target language, we select English data for translation based on four criteria: annotation framework, discourse relation label distribution, genre alignment, and overall dataset size. The detailed correspondences are provided in~\autoref{tab:data_aug}. For each target corpus, we generated augmented data equivalent to approximately 75\% of its original size. This ratio was chosen to enrich the training set without overshadowing the signals from the original in-domain examples. To ensure the quality and relevance of the synthetic data, we implemented a multi-faceted filtering strategy. First, we maintained a label distribution in the augmented set that closely mirrored the original. Second, we aligned the data's genre; for instance, as the German \texttt{deu.rst.pcc} contains mostly editorial texts and news, we primarily drew source material from the `essay' and `news' genres from the English GUM corpus, supplementing it with a small amount of `speech' data to reach the target volume. Finally, we tried to enforce annotation guideline consistency. For example, we observed that German RST annotations do not segment relative clauses, unlike its counterpart in English. Therefore, we excluded any source examples with these incompatible structural patterns from the German augmentation set.

\begin{table}[t]
\centering
\resizebox{\columnwidth}{!}{%
\begin{tabular}{lll}
\toprule
\textbf{Target Corpus} & \textbf{Source Corpus} & \textbf{Selected Source Genres} \\
\midrule
ces.rst.crdt       & eng.erst.gum         & essay, news \\
deu.pdtb.pcc       & eng.pdtb.gum         & essay, news, speech \\
deu.rst.pcc        & eng.erst.gum         & essay, news, speech \\
eus.rst.ert        & eng.erst.gum         & textbook, academic \\
fra.rst.prstc      & eng.erst.gum         & news, academic \\
nld.rst.nldt       & eng.rst.(oll, sts)    & bio, news, letter \\
fas.rst.prstc      & eng.rst.rstdt        & all \\
\bottomrule
\end{tabular}
}
\caption{Source–Target Genre \& Framework Alignment for Translation-Based Data Augmentation}
\vspace{-6pt}
\label{tab:data_aug}
\end{table}

\subsection{Pruned Qwen3-4B Decoder Only}
\label{sec:decoder-only}
For our decoder-only approach, we frame discourse relation classification as a generative task. Specifically, we feed a prompt to a decoder-only model, instructing it to directly select the correct label from a predefined label set included within the prompt itself.
We employ the Qwen3-4B model~\footnote{\url{https://huggingface.co/Qwen/Qwen3-4B}} ~\citep{yang2025qwen3technicalreport}, chosen for its strong multilingual capabilities, supporting over 100 languages and dialects, which aligns well with the multilingual classification task. We apply supervised fine-tuning with instruction-style prompts to improve its task-specific performance.

\paragraph{Pruning} The public Qwen3-4B model originally contains 4.02 billion parameters, slightly exceeding the shared task's 4B parameter limit. To address this, we adopt a pruning strategy based on layer removal as proposed by \citet{men2024shortgptlayerslargelanguage},  which identifies redundant layers by measuring the similarity between their input and output representations. We determined that removing a single, most redundant layer was sufficient to meet the parameter requirement. After fine-tuning, the resulting pruned model~\footnote{\url{https://huggingface.co/JuNymphea/Georgetown-qwen3-4B-pruned-for-disrpt2025}} achieved performance on par with its unpruned counterpart.

\begin{figure*}
    \centering
    \includegraphics[width=0.9\textwidth]
    {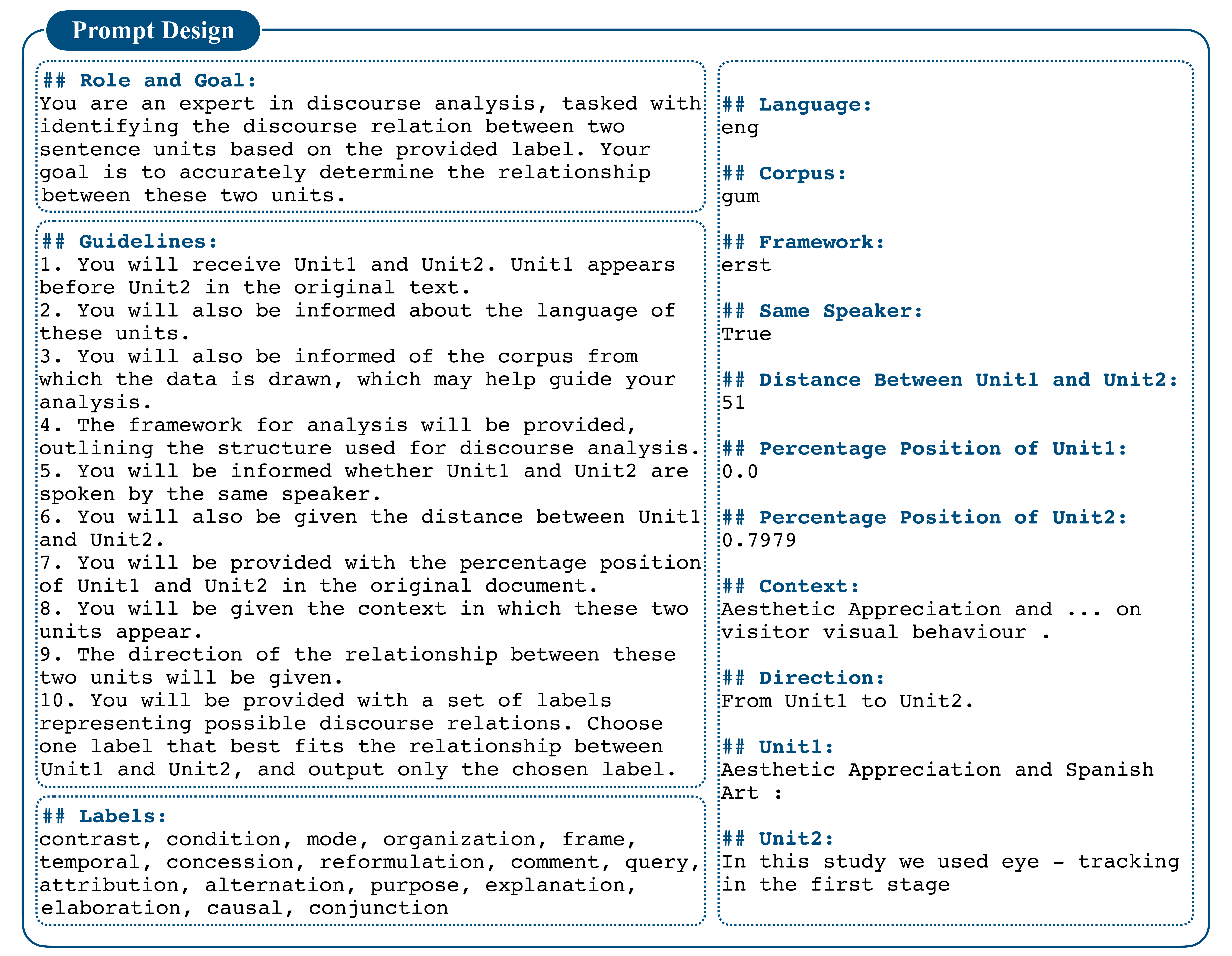}
    \caption{Illustration of the Verbose Instructional Prompt used in Qwen3-4B experiments.}
    \label{fig:prompt}
\end{figure*}

\paragraph{Supervised Fine-Tuning} Our methodology involves full-parameter supervised fine-tuning of the model on the task-specific dataset, which is reformulated into an instruction-style prompt format. Each instruction is enriched with a comprehensive set of features, including LCFs, direction, context, and selected DiscoDisco features (e.g., same speaker, position), as detailed in Section~\ref{sec:features}. The context is constructed from the sentence immediately preceding the first argument, the sentence(s) containing both arguments, and the sentence immediately following the second argument. We experimented with two distinct styles for prompt design: Verbose Instructional Prompt and a Structured Templated Prompt. The verbose prompt, illustrated in~\autoref{fig:prompt}, uses natural language to explicitly define the model's role, the task objective, the various input components, and decision-making guidelines. This contrasts with the structured prompt, which organizes the raw inputs into a compact, delimiter-separated format (e.g., ... \( \text{\$\$ Arg1 \$\$} \) \( > \) \text{\#\# Arg2 \#\#} ...), resembling the input format for encoder models (Section~\ref{sec:encoder}). Although a large part of the verbose instructions is repeated identically in all samples, and may therefore be considered redundant, our experiments consistently showed that the verbose instructional prompts yielded superior performance, improving model accuracy by approximately 1–2\% compared to the more concise, structured variants.


\subsection{mT5 Encoder}
\label{sec:encoder}

We also experimented with mT5 \cite{xue-etal-2021-mt5}, a multilingual T5 variant pretrained on a Common Crawl-based corpus covering 101 languages. Specifically, we selected the mT5‑XL variant, comprising 3.7 billion parameters, which comes closest to the shared task 4B limit. For our classification task, we used only the encoder and added a classification head. We explored two strategies for incorporating metadata and discourse features: (1) encoding them as special input tokens, and (2) using separate embedding layers concatenated with the encoder input.

\paragraph{Feature Injection via Input Tokens}
We prepended LCF features as special input tokens (e.g., \texttt{LANG\_eng, FW\_erst, CORP\_gum}), so that the model can incorporate this metadata directly in its tokenized input sequence. Since mT5 is trained to interpret prompt-like text, this approach naturally lets the model condition on task-specific context. This makes metadata injection especially effective in our setting, where the goal is to classify discourse relations across varied domains and annotation schemes.

In addition to metadata, we also applied this injection strategy to categorical features from the DiscoDisco feature set \cite{gessler-etal-2021-discodisco}, which capture properties such as whether the units are full sentences, whether the relation is discontinuous, and whether the two units share the same speaker. We encoded these features as explicit key-value tokens, for example \texttt{IS\_SENTENCE\_1}, \texttt{DISCONTINUOUS\_0}, and \texttt{SAME\_SPEAKER\_1}. This design differs from the method of \citet{metheniti2024feature}, who append only raw feature values (e.g., \texttt{0.3}, \texttt{0.5}) in a fixed order, with each position implicitly corresponding to a particular feature. While their strategy reduces vocabulary size, it ties interpretation to positional indexing, making it less robust to reordering. By contrast, our approach makes the semantics of each feature explicit and order-independent, which aligns more naturally with mT5’s training paradigm and improves interpretability.

Furthermore, we implemented pseudo-directional features from DiscoDisco. Specifically, for relations labeled as left-to-right (1>2), we inserted direction using the tokens \texttt{\}} and \texttt{>} before and after the first argument span to signal directional flow. For right-to-left (1<2) relations, the inverse markers were used. These directional cues are lightweight but informative, and help disambiguate argument structure across instances, especially in genres with flexible syntax or conversational turn-taking.

The resulting input sequence is organized as follows: metadata (LCF features), followed by categorical DiscoDisco features, and finally the target argument span:
\ex.
\begin{quote}
\small
\texttt{LANG\_eng FW\_erst CORP\_gum [SEP] IS\_SENTENCE\_1 DISCONTINUOUS\_0\linebreak
SAME\_SPEAKER\_1 GENRE\_academic [SEP] \} Aesthetic Appreciation\linebreak
and Spanish Art: > Arg2: In this study we used eye-tracking in the first stage}
\end{quote}

\paragraph{Feature Embedding(s)}
We hypothesized that treating argument spans (arg1 and arg2) together as a single sequence separated by a special token [SEP] might better leverage mT5’s native relative positional embeddings and attention dynamics. Meanwhile, surrounding context (Pre/Post) and metadata features were proposed to be embedded separately, since context can be long and sparse, which may attenuate positional signal strength if concatenated directly with argument spans. Thus, our proposed embedding schema is:

\ex. \texttt{Concat(Embed(meta + features) + Embed(pre-context) + Embed(arg1 [SEP] arg2) + Embed(post-context))}

This setup preserves the association between argument spans, while mitigating positional confusion or noise from long context sequences.

Exploratory experiments on a smaller development subset indicated that this structure offered modest conceptual clarity but did not substantially outperform simply prepending LCF and categorical DiscoDisco features as special tokens. 

\section{Results}

\subsection{Performance Comparison of Encoder-Only and Decoder-Only Model}

\autoref{tab:res_decoder_encoder} reports the test scores of the encoder-only and decoder-only models across all 38 corpora.

\begin{table}[h!tb]
\small
\centering
\begin{tabular}{lcc}
\toprule
\textbf{Corpus} & \textbf{Decoder (DeDisCo)} & \textbf{Encoder}\\
\midrule
ces.rst.crdt & 52.70 & 51.35\\
deu.pdtb.pcc & 67.01 & 56.19\\
deu.rst.pcc & 67.03 & 49.82\\
eng.dep.covdtb & 68.21 & 73.05\\
eng.dep.scidtb & 83.66 & 79.58\\
eng.erst.gentle & 67.08 & 61.29\\
eng.erst.gum & 73.45 & 62.98\\
eng.pdtb.gentle & 67.94 & 61.07\\
eng.pdtb.gum & 71.39 & 65.20\\
eng.pdtb.pdtb & 83.77 & 77.32\\
eng.pdtb.tedm & 71.79 & 61.54\\
eng.rst.oll & 62.73 & 51.66\\
eng.rst.rstdt & 73.27 & 62.60\\
eng.rst.sts & 58.54 & 49.39\\
eng.rst.umuc & 67.36 & 59.09\\
eng.sdrt.msdc & 90.00 & 84.11\\
eng.sdrt.stac & 75.80 & 65.96\\
eus.rst.ert & 54.64 & 55.67\\
fas.rst.prstc & 60.47 & 57.77\\
fra.sdrt.annodis & 60.39 & 52.82\\
ita.pdtb.luna & 70.13 & 66.13\\
nld.rst.nldt & 68.62 & 53.85\\
pcm.pdtb.disconaija & 59.39 & 41.40\\
pol.iso.pdc & 74.02 & 55.05\\
por.pdtb.crpc & 79.17 & 75.64\\
por.pdtb.tedm & 68.41 & 64.84\\
por.rst.cstn & 70.22 & 69.85\\
rus.rst.rrt & 74.85 & 68.95\\
spa.rst.rststb & 69.72 & 64.55\\
spa.rst.sctb & 83.02 & 76.73\\
tha.pdtb.tdtb & 96.73 & 96.80\\
tur.pdtb.tdb & 64.13 & 65.08\\
tur.pdtb.tedm & 59.23 & 54.55\\
zho.dep.scidtb & 80.00 & 68.37\\
zho.pdtb.cdtb & 88.65 & 86.54\\
zho.pdtb.ted & 75.49 & 66.24\\
zho.rst.gcdt & 75.55 & 65.37\\
zho.rst.sctb & 74.21 & 66.67\\
\midrule
\textbf{Macro Average} & 71.28 & 64.34\\
\textbf{Micro Average} & 76.13 & 69.74\\
\bottomrule
\end{tabular}
\caption{Accuracy of encoder-only and decoder-only models on the test sets of 38 corpora. We use corpus codes for simplicity (check \autoref{tab:corpora-list} in \cref{sec:appendix-corpora} for language and framework information), with accuracy scores reported separately for the decoder-only and encoder-only models. Macro and micro averages are reported at the bottom.}
\label{tab:res_decoder_encoder}
\vspace{-6pt}
\end{table}

We note several coarse observations at the outset: First, the decoder model outperforms the encoder model in all datasets except four: \texttt{eng.dep.covdtb, eus.rst.ert, tur.pdtb.tdb, tha.pdtb.tdtb}. However, the difference is minor (around 1 accuracy point or less) in all but \texttt{eng.dep.covdtb}, which is a `test-only' dataset, meaning the models have seen no data of the same kind -- although a few other datasets are also test-only, they do have `related' datasets (\texttt{eng.erst.gentle}, \citealt{aoyama-etal-2023-gentle}, is modeled on \texttt{eng.erst.gum}, and the \texttt{*.tedm} datasets closely follow recent versions of PDTB, \citealt{prasad-etal-2018-discourse}). Aside from this unique property, we are unsure what sets this dataset apart. 

By contrast, in the most extreme case the decoder achieves a 19 gain compared to the encoder on the Polish \texttt{pol.iso.pdc}. We can rule out data contamination with gold data as a reason, since the data was released in the \textit{surprise} test set and was reported to be annotated very recently. 

Beyond individual dataset differences, we observed a broader trend with respect to data scale. In lower-resource settings, such as Czech (14.6K tokens), Dutch (24.9K), and French (32.7K), the decoder model consistently shows a substantial advantage over the encoder, with accuracy gains exceeding 10 points in some cases. These datasets have limited supervision and lack related training corpora, making them particularly reliant on pretrained representations. The decoder’s autoregressive architecture and larger capacity appear to enable better generalization under these conditions. On the other hand, in larger datasets such as Thai (256K tokens), Turkish (496K), and English PDTB (1.17M), the performance gap narrows. In fact, the encoder slightly outperforms the decoder on Thai PDTB, suggesting that when sufficient labeled data is available, the simpler encoder-only setup can be just as effective, if not more so. 

We also observed differences in how the two architectures respond to feature integration. LCF and DiscoDisco features consistently improved the encoder model, but in many cases degraded performance for the decoder. This suggests that encoder‑only models can more effectively leverage categorical metadata and structural cues as additional signals, whereas decoder‑only models are more sensitive to such injections. In contrast, extending the context window benefited the decoder but often harmed the encoder.

\subsection{Decoder Model Ablation Tests}


To assess the contribution of each additional features, we conduct an ablation study on the decoder-only model. In each experiment, one specific feature is removed from the input to evaluate its impact on performance. Detailed results are presented in \cref{sec:appendix-ablation} (~\autoref{tab:res_ablation}).

We find that direction is the most influential feature. When direction information is removed, 32 out of the 38 corpora experience a drop in accuracy of more than 4\%, and 12 corpora suffer a decrease of over 10\%. By contrast, for \texttt{tha.pdtb.tdtb}, direction contributes only marginal gains, and for \texttt{tur.pdtb.tedm}, it even leads to a performance drop. This is likely because Turkish is a free-word-order left‑branching language, where in most subordinate constructions the clause with the connective precedes the main clause, causing direction features based on linear order to introduce noise rather than useful signal \cite{zeyrek2009annotating}. 

The second most impactful feature is context. Although prior work~\citep{judge2024analysissententialneighborsimplicit} suggests that adding context may not always be beneficial, we observe that context is a highly effective input for the decoder-only model. However, its impact varies significantly across corpora. For instance, \texttt{deu.rst.pcc} and \texttt{pcm.pdtb.disconaija} see gains exceeding 15\%, and even up to 20\% in \texttt{pol.iso.pdc}. For most corpora, the improvement is around 3\%, though performance actually deteriorates on \texttt{tur.pdtb.tdb} and \texttt{ita.pdtb.luna}.

By comparison, LCF features tend to have smaller or even negative effects across many corpora. Nevertheless, they yield notable improvements (greater than 5\%) for \texttt{deu.rst.pcc} and \texttt{zho.rst.sctb}. 

A surprising finding emerges regarding data augmentation: it does not always improve performance on the target corpus. For example, For example, we observe gains on five corpora, except for \texttt{ces.rst.crdt} and \texttt{deu.pdtb.pcc}. On the other hand, for the source corpora, all five English datasets show consistent improvements from data augmentation, despite the source samples for the augmented entries also existing in the English training.

As for the DiscoDisco features, their overall impact was less pronounced across most corpora. The most notable exception was \texttt{eng.dep.covdtb}. For this unique, test-only dataset with no in-domain training data, removing these features surprisingly boosted performance by nearly 5\%, suggesting they may introduce counterproductive noise.

\begin{figure}
  \centering
  \includegraphics[width=\linewidth]{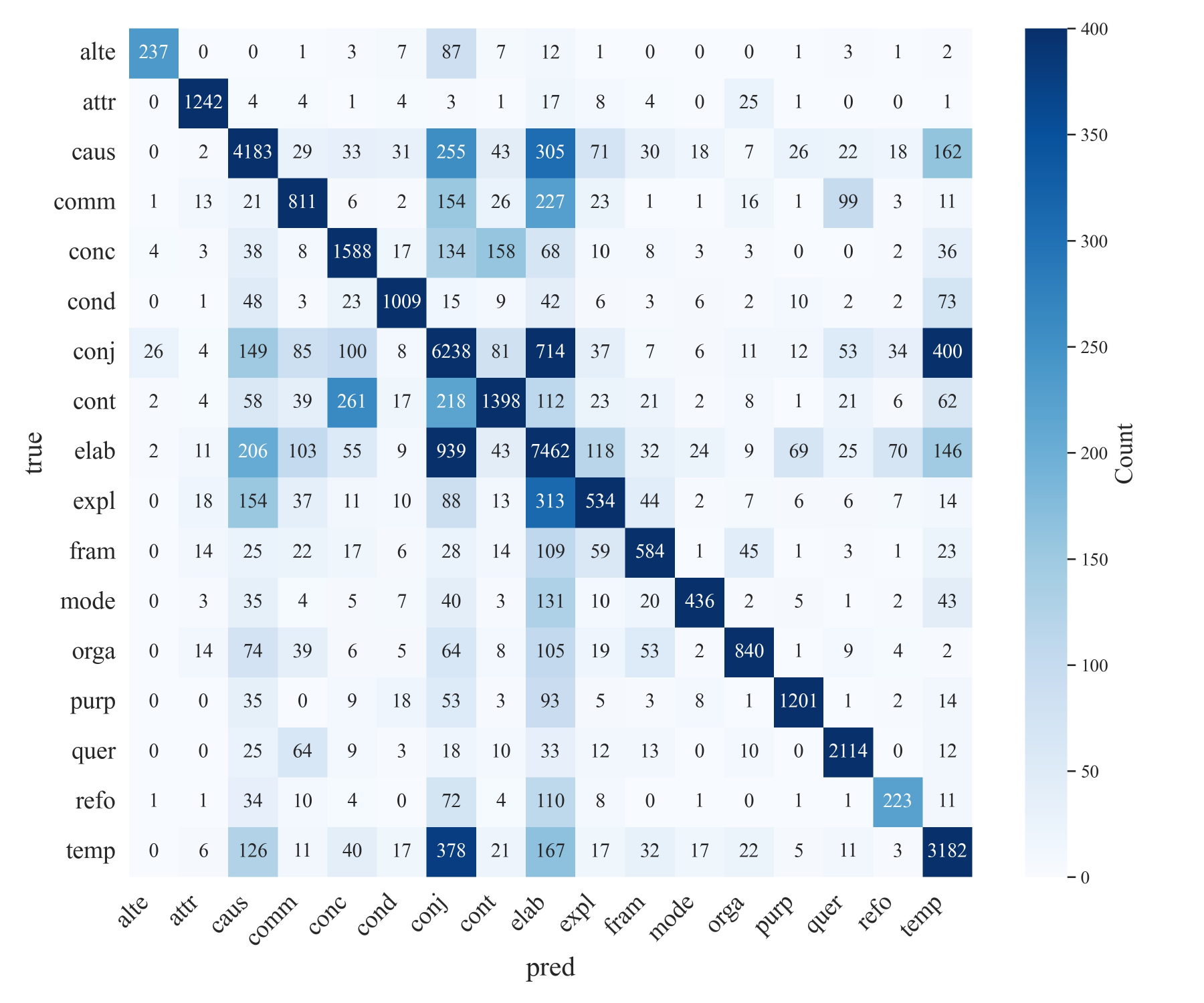}
  \caption{Confusion matrix over the entire dataset.}
  \label{fig:confmat_all}
  \vspace{-6pt}
\end{figure}

\begin{figure*}%
    \centering
    \begin{subfigure}{0.48\textwidth}
    \includegraphics[width=\linewidth]{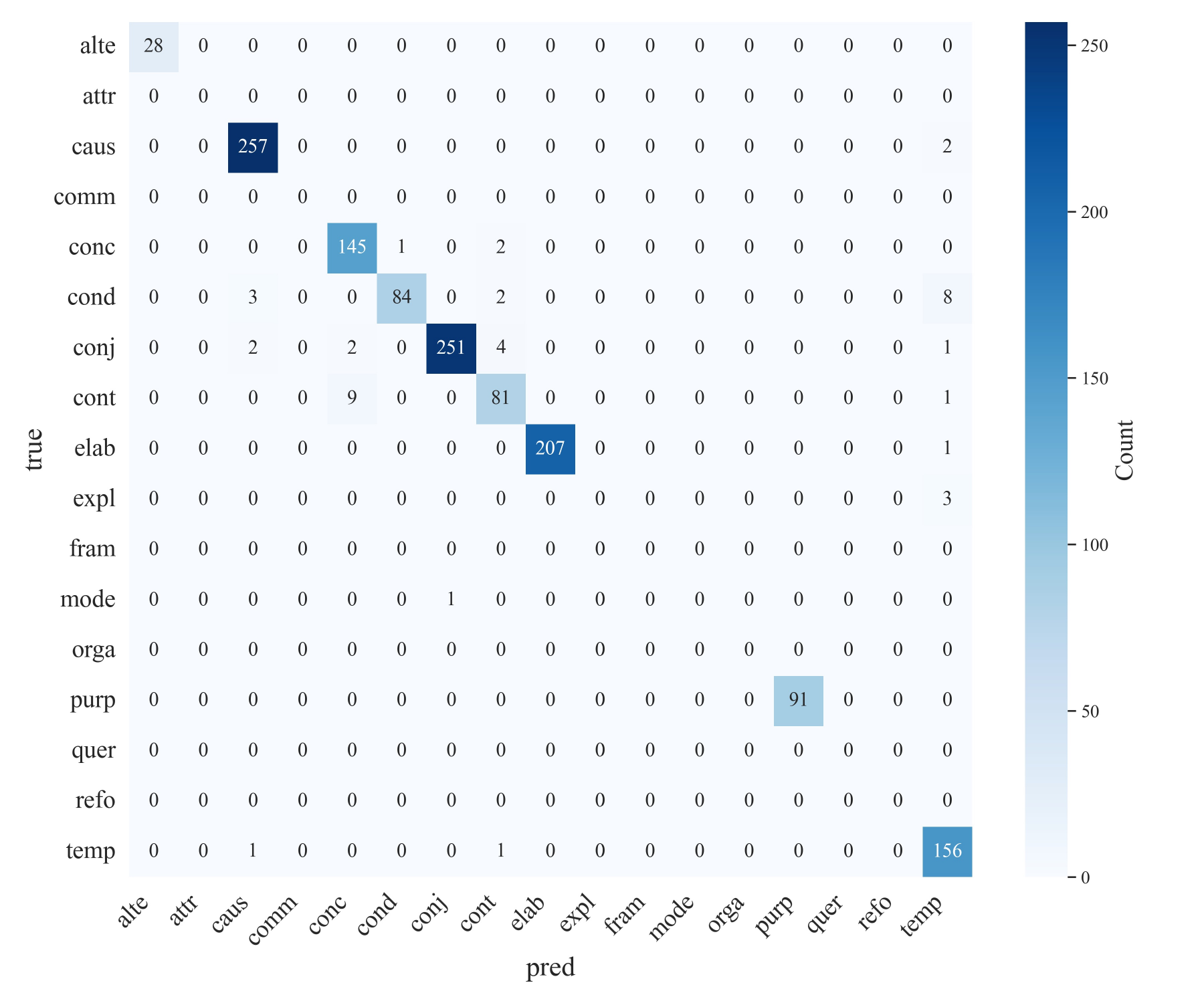}
    \caption{tha.pdtb.tdtb}
   \end{subfigure}
    \begin{subfigure}{0.48\textwidth}
    \includegraphics[width=\linewidth]{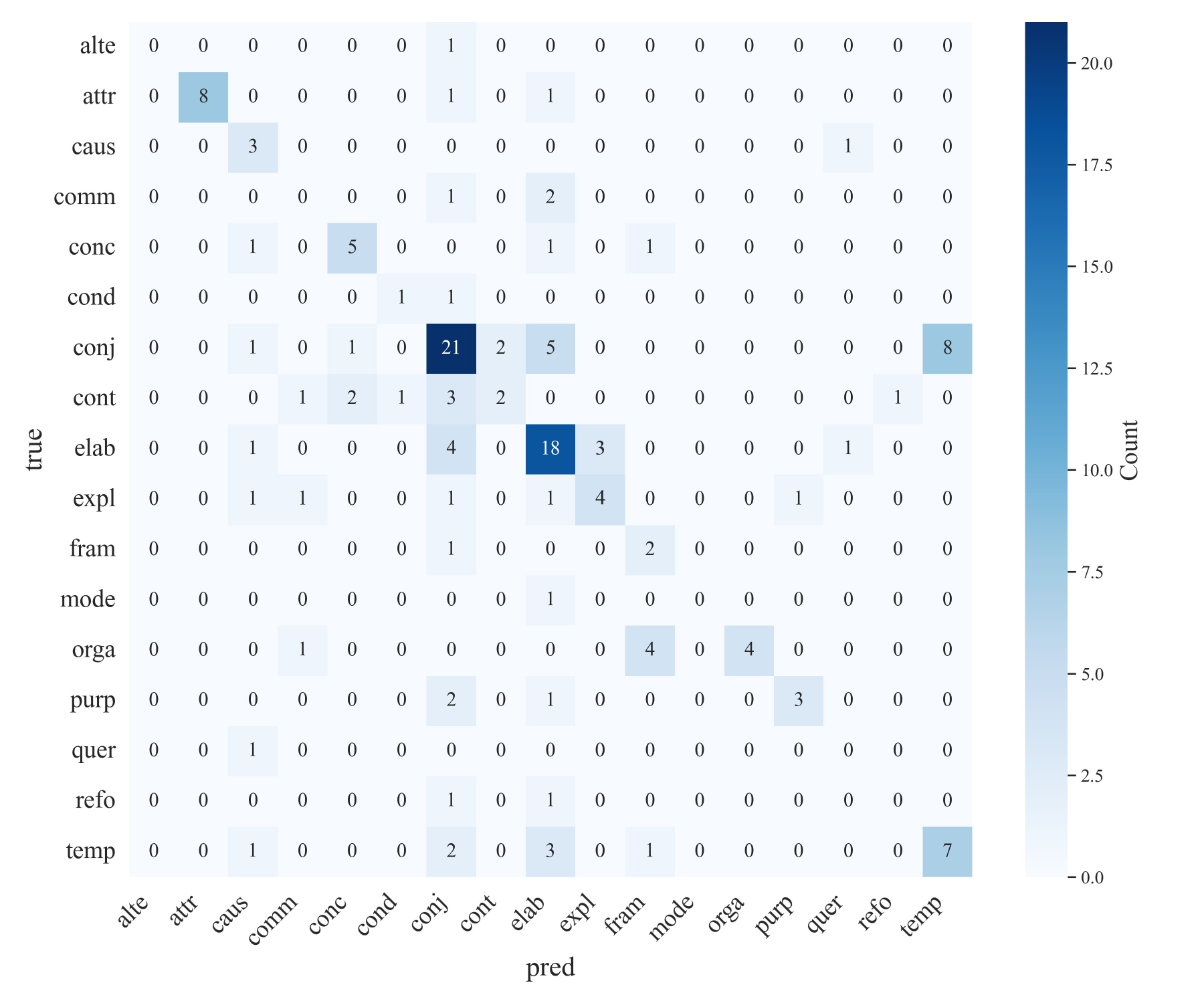}
    \caption{ces.rst.crdt}
\end{subfigure}
    \caption{Confusion Matrices for Common Relations in the Highest and Lowest Scoring Datasets.}%
    \label{fig:highest_lowest_datasets}%
\end{figure*}

\section{Error analysis}


\paragraph{Strengths and Confusions at the Relation Level}
The confusion matrix in \autoref{fig:confmat_all} indicates very high performance on key discourse relations including \textsc{elaboration}, \textsc{conjunction}, \textsc{causal}, and \textsc{temporal}, demonstrating that the model reliably classifies these central categories. However, we observed minor yet notable confusion between \textsc{elaboration} and \textsc{conjunction}, and this might because that \textsc{elaboration} and \textsc{conjunction} are frequently conflated due to their semantic and structural similarity, especially when explicit lexical signals are absent or ambiguous, which is often the case for \textsc{elaboration}, the most over-predicted label in the dataset. Teasing apart \textsc{conjunction} and \textsc{temporal} is also challenging, especially in cases where consecutive events are not signaled explicitly, but implicitly form a temporal succession relation.

\paragraph{Strengths and Confusions at the Corpus Level}
\autoref{fig:highest_lowest_datasets} shows the confusion matrices for the highest- and lowest-scoring datasets, \texttt{tha.pdtb.tdtb} (TDTB) and \texttt{ces.rst.crdt} (CRDT). In CRDT, the model frequently defaults to the majority relation \textsc{conjunction}, reflecting strong over-prediction of common classes. The label distribution is also highly imbalanced: \textsc{conjunction} and \textsc{elaboration} dominate the dataset, while many relations (e.g., \textsc{alternative}, \textsc{reformulation}, \textsc{mode}) appear only rarely, making them difficult for the model to learn.

Relations marked by overt lexical cues, such as \textsc{conjunction}, achieve high accuracy in both datasets. Performance in TDTB is further aided by the fact that this dataset covers only 12/17 possible labels, substantially reducing the possibilities for confusion compared to CRDT, which covers the full set of 17 labels.

\section{Conclusion}
We present \textbf{DeDisCo}, a decoder-only model, using a pruned Qwen3-4B basis, for the multilingual discourse relation classification task in the DISRPT 2025 Shared Task. Our system DeDisCo leverages supervised fine-tuning together with rich features, including metadata and instance-level cues such as unit distance, document position, and gold speaker information. We also incorporated augmented datasets to improve coverage for low-resource languages.

Our results suggest that decoder-only architectures are effective for this task, as their structure allows the model to integrate diverse sources of information (e.g., metadata, features, and context) within a unified text stream. Carefully designed instruction templates and feature injection further improve generalization, and natural prompt styles are helpful despite textual redundancy, even in a full fine-tuning setup. This enables the model to condition on a broad range of cues when making predictions, while adhering to a format the model is familiar with from its initial supervised fine-tuning. Error analysis highlights class imbalance as a persistent challenge, often leading to over-prediction of majority classes. Nonetheless, augmented data yielded measurable gains for low-resource languages, underscoring the importance of data enrichment strategies in this task.

\bibliography{custom, language_resources}

\appendix

\section{Data}

We train and evaluate our models using all the datasets provided by the shared task organizers.\footnote{\url{https://github.com/disrpt/sharedtask2025}}
In total, the benchmark is composed of $39$ datasets, covering $13$ languages and $6$ frameworks.
These datasets were obtained from the following corpora: 
the Czech RST Discourse Treebank 1.0 \cite{11234/1-5174},
the Potsdam Commentary Corpus \cite{StedeNeumann2014,bourgonje-stede-lrec2020},
the COVID-19 Discourse Dependency Treebank \cite{nishida-matsumoto-2022-domain},
the Discourse Dependency TreeBank for Scientific Abstracts \cite{yang-li-2018-scidtb,yi-etal-2021-unifying,cheng-li-2019-zero},
the Genre Tests for Linguistic Evaluation corpus \cite{aoyama-etal-2023-gentle},
the Georgetown University Multilayer corpus \cite{Zeldes2017},
the RST Discourse Treebank \cite{carlson-etal-2001-building},
the Science, Technology, and Society corpus  \cite{potter2008interactional},
the University of Potsdam Multilayer UNSC Corpus \cite{zaczynska-stede-2024-rhetorical},
the Minecraft Structured Dialogue Corpus \cite{thompson-etal-2024-discourse},
the Strategic Conversations corpus \cite{asher-etal-2016-discourse},
the Basque RST Treebank \cite{IruskietaAranzabeIlarrazaEtAl2013},
the Persian RST Corpus \cite{shahmohammadi2021persian},
the ANNOtation DIScursive corpus \cite{afantenos-etal-2012-empirical},
the SUMM-RE corpus \cite{hunter-etal-2024-meeting,hunter_summre_2025},
the Dutch Discourse Treebank \cite{redeker-etal-2012-multi},
the Polish Discourse Corpus \cite{ogr:etal:24,lrec:coling:24},
the Cross-document Structure Theory News Corpus \cite{CardosoMazieroRosarioCastroJorgeEtAl2011},
the Russian RST Treebank \cite{toldova-etal-2017-rhetorical},
the RST Spanish Treebank \cite{da-cunha-etal-2011-development},
the RST Spanish-Chinese Treebank \cite{cao-etal-2018-rst},
the Georgetown Chinese Discourse Treebank \cite{peng_gcdt_2022,peng_chinese_2022},
the DiscoNaija corpus \cite{scholman2025disconaija},
the Penn Discourse Treebank \cite{prasad-etal-2014-reflections,PDTB3-Annotation-Manual},
the TED-Multilingual Discourse Bank (English) \cite{zeyrek-etal-2018-multilingual,zeyrek2019ted},
the LUNA Corpus Discourse Data Set \cite{tonelli-etal-2010-annotation,RiccardiStepanovChowdhury2016},
the Portuguese Discourse Bank  \cite{CRPC-DB-Portuguese,genereux-etal-2012-introducing},
the Thai Discourse Treebank \cite{thai_prasertsom_2024},
the Turkish Discourse Bank \cite{zeyrek-webber-2008-discourse,zeyrek-kurfali-2017-tdb},
and
the Chinese Discourse Treebank \cite{Zhou2014}.

\section{Feature Utilization in Decoder and Encoder Models}
\label{sec:appendix-feature}

\begin{table}[h]
\centering
\begin{tabular}{lcc}
\hline
\textbf{Feature Group} & \textbf{Decoder} & \textbf{Encoder} \\
\hline
LCF features              & + & + \\
DiscoDisco features       & + & + \\
Direction                 & + & + \\
Context (window size)     & + & – \\
Augmented dataset         & + & + \\
\hline
\end{tabular}
\caption{Feature groups and dataset augmentation used in decoder vs.\ encoder models (“+” = included, “–” = excluded).}
\label{tab:overview_features}
\end{table}

\begin{table}[h]
\centering
\begin{tabular}{lcc}
\hline
\textbf{DiscoDisco Feature} & \textbf{Decoder} & \textbf{Encoder} \\
\hline
Genre             & – & + \\
Children          & – & – \\
Discontinuous     & – & + \\
Is Sentence       & – & + \\
Length Ratio      & – & – \\
Same Speaker      & + & + \\
Document Length   & – & – \\
Position          & + & – \\
Distance          & + & – \\
Lexical Overlap   & – & – \\
\hline
\end{tabular}
\caption{Detailed inclusion (“+”) or exclusion (“–”) of all DiscoDisco features for each model.}
\label{tab:discodisco_breakdown}
\end{table}

\section{Experimental Setup of Decoder}
\label{sec:appendix-exps}

The model was trained on four NVIDIA H100 GPUs. Training one epoch took approximately three hours, and evaluation on the test sets required an additional one and a half hours. With a per-device batch size of 1, and using 16 gradient accumulation steps, the effective batch size was 64.

On rare occasions, the generative model produced outputs that were not part of the predefined set of valid labels. In such cases, our evaluation script replaced the output with a randomly selected valid label. This was extremely infrequent, occurring fewer than five times across all evaluations.

\section{Corpora from DISRPT 2025 Shared Task}
\label{sec:appendix-corpora}
\begin{table}[H]
    \centering
    \resizebox{\columnwidth}{!}{ 
    \begin{tabular}{lll}
    \toprule
    Language  & Framework & Corpus \\
    Name (Code) & & LFC Short Code\\
    \midrule
    Czech (ces) & RST &ces.rst.crdt \\
    Standard German (deu) & PDTB &deu.pdtb.pcc \\
    Standard German (deu) & RST &deu.rst.pcc \\
    English (eng) & DEP & eng.dep.covdtb \\
    English (eng) & DEP & eng.dep.scidtb \\
    English (eng) & eRST & eng.erst.gentle \\
    English (eng) & eRST & eng.erst.gum \\
    English (eng) & PDTB & eng.pdtb.gentle \\
    English (eng) & PDTB & eng.pdtb.gum \\
    English (eng) & PDTB & eng.pdtb.pdtb \\
    English (eng) & PDTB & eng.pdtb.tedm \\
    English (eng) & RST & eng.rst.oll \\
    English (eng) & RST & eng.rst.rstdt \\
    English (eng) & RST & eng.rst.sts \\
    English (eng) & RST & eng.rst.umuc \\
    English (eng) & SDRT  &eng.sdrt.msdc \\
    English (eng) & SDRT & eng.sdrt.stac \\
    Basque (eus) & RST & eus.rst.ert \\
    Persian (fas) & RST &f as.rst.prstc \\
    French (fra) & SDRT & fra.sdrt.annodis \\
    French (fra) & SDRT & fra.sdrt.summre$^\prime$\\
    Italian (ita) & PDTB & ita.pdtb.luna \\
    Dutch (nld) & RST & nld.rst.nldt \\
    Nigerian Pidgin (pcm) & PDTB & pcm.pdtb.disconaija \\
    Polish (pol) & ISO & pol.iso.pdc \\
    Portuguese (por) & PDTB & por.pdtb.crpc \\
    Portuguese (por) & PDTB & por.pdtb.tedm \\
    Portuguese (por) & RST & por.rst.cstn \\
    Russian (rus) & RST & rus.rst.rrt \\
    Spanish (spa) & RST & spa.rst.rststb \\
    Spanish (spa) & RST & spa.rst.sctb \\
    Thai (tha) & PDTB & tha.pdtb.tdtb \\
    Turkey (tur) & PDTB & tur.pdtb.tdb \\
    Turkey (tur) & PDTB & tur.pdtb.tedm \\
    Chinese (zho) & DEP & zho.dep.scidtb \\
    Chinese (zho) & PDTB & zho.pdtb.cdtb \\
    Chinese (zho) & PDTB & zho.pdtb.ted \\
    Chinese (zho) & RST & zho.rst.gcdt \\
    Chinese (zho) & RST & zho.rst.sctb \\    
\bottomrule
\end{tabular}
}
    \caption{Collation of the data in DISRPT 2025 Shared task with the corresponding Language and Framework for reference. The frameworks include RST~\cite{mann1988rhetorical}, PDTB~\cite{marcu2004phrase}, DEP~\cite{stede2016parallel}, SDRT~\cite{lascarides2007segmented}, eRST~\cite{zeldes-etal-2025-erst}, and ISO 24617-8 \cite{bunt2016iso, tomaszewska-etal-2024-iso}. The languages are sorted based on the language code (ISO 639-3\footnotemark). The corpus \texttt{fra.sdrt.summre} marked with $^\prime$ is not part of the relation classification task.}
    \label{tab:corpora-list}
\end{table}

\footnotetext{\url{https://www.iso.org/iso-639-language-code}}

\clearpage
\section{Ablation Test Results of Decoder}
\label{sec:appendix-ablation}
\begin{table}[H]
\begin{minipage}{\textwidth}
\resizebox{\textwidth}{!}{%
\begin{tabular}{lccccccccccc}
\toprule
\textbf{Corpus} & \textbf{Decoder} & \multicolumn{2}{c}{\textbf{w/o LCF}} & \multicolumn{2}{c}{\textbf{w/o DiscoDisco}} & \multicolumn{2}{c}{\textbf{w/o Direction}} & \multicolumn{2}{c}{\textbf{w/o Context}} & \multicolumn{2}{c}{\textbf{w/o Aug}} \\
&&\textit{abs.}&\textit{gain}&\textit{abs.}&\textit{gain}&\textit{abs.}&\textit{gain}&\textit{abs.}&\textit{gain}&\textit{abs.}&\textit{gain}\\
\midrule
ces.rst.crdt & 52.70 & 54.73 & -2.03 & 55.41 & -2.71 & 46.62 & 6.08 & 50.00 & 2.70 & 57.43 & -4.73\\
deu.pdtb.pcc & 67.01 & 63.92 & 3.09 & 65.98 & 1.03 & 62.37 & 4.64 & 62.37 & 4.64 & 69.07 & -2.06\\
deu.rst.pcc & 67.03 & 59.71 & 7.32 & 64.84 & 2.19 & 48.72 & 18.31 & 51.28 & 15.75 & 64.10 & 2.93\\
eng.dep.covdtb & 68.21 & 70.46 & -2.25 & 73.16 & -4.95 & 57.39 & 10.82 & 71.04 & -2.83 & 71.81 & -3.6\\
eng.dep.scidtb & 83.66 & 84.29 & -0.63 & 83.87 & -0.21 & 82.57 & 1.09 & 81.05 & 2.61 & 83.56 & 0.1\\
eng.erst.gentle & 67.08 & 62.93 & 4.15 & 65.99 & 1.09 & 60.27 & 6.81 & 64.62 & 2.46 & 67.32 & -0.24\\
eng.erst.gum & 73.45 & 71.29 & 2.16 & 73.11 & 0.34 & 64.72 & 8.73 & 66.84 & 6.61 & 72.41 & 1.04\\
eng.pdtb.gentle & 67.94 & 65.27 & 2.67 & 67.05 & 0.89 & 63.61 & 4.33 & 67.43 & 0.51 & 66.79 & 1.15\\
eng.pdtb.gum & 71.39 & 68.00 & 3.39 & 70.86 & 0.53 & 65.38 & 6.01 & 68.47 & 2.92 & 70.8 & 0.59\\
eng.pdtb.pdtb & 83.77 & 83.19 & 0.58 & 84.07 & -0.30 & 73.22 & 10.55 & 82.61 & 1.16 & 83.80 & -0.03\\
eng.pdtb.tedm & 71.79 & 68.95 & 2.84 & 71.23 & 0.56 & 64.67 & 7.12 & 68.09 & 3.70 & 69.52 & 2.27\\
eng.rst.oll & 62.73 & 60.89 & 1.84 & 60.52 & 2.21 & 49.45 & 13.28 & 59.41 & 3.32 & 59.04 & 3.69\\
eng.rst.rstdt & 73.27 & 67.38 & 5.89 & 73.41 & -0.14 & 68.54 & 4.73 & 69.51 & 3.76 & 72.99 & 0.28\\
eng.rst.sts & 58.54 & 55.49 & 3.05 & 56.71 & 1.83 & 45.43 & 13.11 & 50.00 & 8.54 & 54.88 & 3.66\\
eng.rst.umuc & 67.36 & 66.53 & 0.83 & 66.53 & 0.83 & 61.16 & 6.2 & 61.16 & 6.20 & 62.4 & 4.96\\
eng.sdrt.msdc & 90.00 & 89.75 & 0.25 & 90.03 & -0.03 & 88.82 & 1.18 & 86.14 & 3.86 & 89.08 & 0.92\\
eng.sdrt.stac & 75.80 & 76.33 & -0.53 & 75.98 & -0.18 & 70.92 & 4.88 & 70.66 & 5.14 & 74.73 & 1.07\\
eus.rst.ert & 54.64 & 56.49 & -1.85 & 51.96 & 2.68 & 43.30 & 11.34 & 46.60 & 8.04 & 52.16 & 2.48\\
fas.rst.prstc & 60.47 & 59.12 & 1.35 & 59.12 & 1.35 & 51.52 & 8.95 & 50.84 & 9.63 & 59.63 & 0.84\\
fra.sdrt.annodis & 60.39 & 56.04 & 4.35 & 61.19 & -0.80 & 53.14 & 7.25 & 51.53 & 8.86 & 58.94 & 1.45\\
ita.pdtb.luna & 70.13 & 70.40 & -0.27 & 70.13 & 0 & 61.6 & 8.53 & 70.93 & -0.80 & 72.53 & -2.40\\
nld.rst.nldt & 68.62 & 69.85 & -1.23 & 69.54 & -0.92 & 55.08 & 13.54 & 61.23 & 7.39 & 67.69 & 0.93\\
pcm.pdtb.disconaija & 59.39 & 60.96 & -1.57 & 61.16 & -1.77 & 51.13 & 8.26 & 42.97 & 16.42 & 60.18 & -0.79\\
pol.iso.pdc & 74.02 & 73.08 & 0.94 & 73.62 & 0.40 & 62.99 & 11.03 & 53.97 & 20.05 & 72.14 & 1.88\\
por.pdtb.crpc & 79.17 & 79.09 & 0.08 & 78.85 & 0.32 & 73.48 & 5.69 & 77.48 & 1.69 & 77.72 & 1.45\\
por.pdtb.tedm & 68.41 & 68.68 & -0.27 & 68.41 & 0 & 65.11 & 3.30 & 65.38 & 3.03 & 67.03 & 1.38\\
por.rst.cstn & 70.22 & 70.96 & -0.74 & 71.32 & -1.10 & 69.12 & 1.10 & 70.59 & -0.37 & 71.32 & -1.1\\
rus.rst.rrt & 74.85 & 74.81 & 0.04 & 75.31 & -0.46 & 66.18 & 8.67 & 69.52 & 5.33 & 74.46 & 0.39\\
spa.rst.rststb & 69.72 & 71.83 & -2.11 & 70.66 & -0.94 & 64.55 & 5.17 & 65.96 & 3.76 & 70.42 & -0.70\\
spa.rst.sctb & 83.02 & 77.99 & 5.03 & 80.50 & 2.52 & 70.44 & 12.58 & 76.73 & 6.29 & 86.16 & -3.14\\
tha.pdtb.tdtb & 96.73 & 96.88 & -0.15 & 96.5 & 0.23 & 96.58 & 0.15 & 96.73 & 0 & 96.50 & 0.23\\
tur.pdtb.tdb & 64.13 & 66.03 & -1.90 & 66.75 & -2.62 & 59.86 & 4.27 & 64.61 & -0.48 & 66.75 & -2.62\\
tur.pdtb.tedm & 59.23 & 58.4 & 0.83 & 59.78 & -0.55 & 59.5 & -0.27 & 58.95 & 0.28 & 58.4 & 0.83\\
zho.dep.scidtb & 80.00 & 78.6 & 1.40 & 77.21 & 2.79 & 69.77 & 10.23 & 74.42 & 5.58 & 76.28 & 3.72\\
zho.pdtb.cdtb & 88.65 & 88.52 & 0.13 & 90.5 & -1.85 & 83.91 & 4.74 & 87.34 & 1.31 & 88.79 & -0.14\\
zho.pdtb.ted & 75.49 & 76.09 & -0.60 & 75.86 & -0.37 & 67.97 & 7.52 & 71.95 & 3.54 & 75.79 & -0.30\\
zho.rst.gcdt & 75.55 & 73.66 & 1.89 & 75.13 & 0.42 & 62.96 & 12.59 & 70.51 & 5.04 & 76.71 & -1.16\\
zho.rst.sctb & 74.21 & 67.30 & 6.91 & 73.58 & 0.63 & 62.26 & 11.95 & 70.44 & 3.77 & 71.70 & 2.51\\
\midrule
\textbf{Macro Average} & 71.28 & 70.10 & 1.18 & 71.21 & 0.08 & 63.80 & 7.49 & 66.56 & 4.72 & 70.82 & 0.47\\
\textbf{Micro Average} & 76.13 & 75.15 & 0.98 & 76.38 & -0.25 & 69.53 & 6.60 & 72.24 & 3.89 & 75.86 & 0.27\\
\bottomrule
\end{tabular}
}

\captionof{table}{{Accuracy results of the ablation study on the decoder-only model: next to the scores from Table ~\ref{tab:res_decoder_encoder}, we report scores without LCF features, without DiscoDisco features, without direction, without context and without data augmentation, as well as the “gain” for each (non-ablated score – ablated score). }}
\label{tab:res_ablation}%
\end{minipage}
\end{table}

\end{document}